\def\year{2019}\relax
\documentclass[letterpaper]{article} 
\usepackage{aaai19}  
\usepackage{times}  
\usepackage{helvet}  
\usepackage{courier}  
\usepackage{url}  
\usepackage{graphicx}  
\usepackage{amsfonts}

\usepackage{booktabs}
\usepackage{amssymb}
\usepackage{multirow}
\usepackage{multicol}

\begin{document}
%
\title{MeshNet: Mesh Neural Network for 3D Shape Representation}
\author{
Yutong Feng,\textsuperscript{1}
Yifan Feng,\textsuperscript{2}
Haoxuan You,\textsuperscript{1}
Xibin Zhao\textsuperscript{1}\thanks{Corresponding authors},
Yue Gao\textsuperscript{1}\footnotemark[1]\\
\textsuperscript{1}BNRist, KLISS, School of Software, Tsinghua University, China.\\
\textsuperscript{2}School of Information Science and Engineering, Xiamen University\\
\{feng-yt15, zxb, gaoyue\}@tsinghua.edu.cn,
\{evanfeng97, haoxuanyou\}@gmail.com\\
}
\maketitle

\begin{abstract}
Mesh is an important and powerful type of data for 3D shapes and widely studied in the field of computer vision and computer graphics. Regarding the task of 3D shape representation, there have been extensive research efforts concentrating on how to represent 3D shapes well using volumetric grid, multi-view and point cloud. However, there is little effort on using mesh data in recent years, due to the complexity and irregularity of mesh data. In this paper, we propose a mesh neural network, named MeshNet, to learn 3D shape representation from mesh data. In this method, face-unit and feature splitting are introduced, and a general architecture with available and effective blocks are proposed. In this way, MeshNet is able to solve the complexity and irregularity problem of mesh and conduct 3D shape representation well. We have applied the proposed MeshNet method in the applications of 3D shape classification and retrieval. Experimental results and comparisons with the state-of-the-art methods demonstrate that the proposed MeshNet can achieve satisfying 3D shape classification and retrieval performance, which indicates the effectiveness of the proposed method on 3D shape representation.

\end{abstract}

\section{Introduction}
Three-dimensional (3D) shape representation is one of the most fundamental topics in the field of computer vision and computer graphics. In recent years, with the increasing applications in 3D shapes, extensive efforts \cite{wu20153d,chang2015shapenet} have been concentrated on 3D shape representation and proposed methods are successfully applied for different tasks, such as classification and retrieval.

For 3D shapes, there are several popular types of data, including volumetric grid, multi-view, point cloud and mesh. With the success of deep learning methods in computer vision, many neural network methods have been introduced to conduct 3D shape representation using volumetric grid \cite{wu20153d,maturana2015voxnet}, multi-view \cite{su2015multi} and point cloud \cite{qi2017pointnet}. PointNet \cite{qi2017pointnet} proposes to learn on point cloud directly and solves the disorder problem with per-point Multi-Layer-Perceptron (MLP) and a symmetry function. As shown in Figure 1, although there have been recent successful methods using the types of volumetric grid, multi-view and point cloud, for the mesh data, there are only early methods using handcraft features directly, such as the Spherical Harmonic descriptor (SPH) \cite{kazhdan2003rotation}, which limits the applications of mesh data.

\begin{figure}[!tbp]
    \centering
    \includegraphics[width=3.6in]{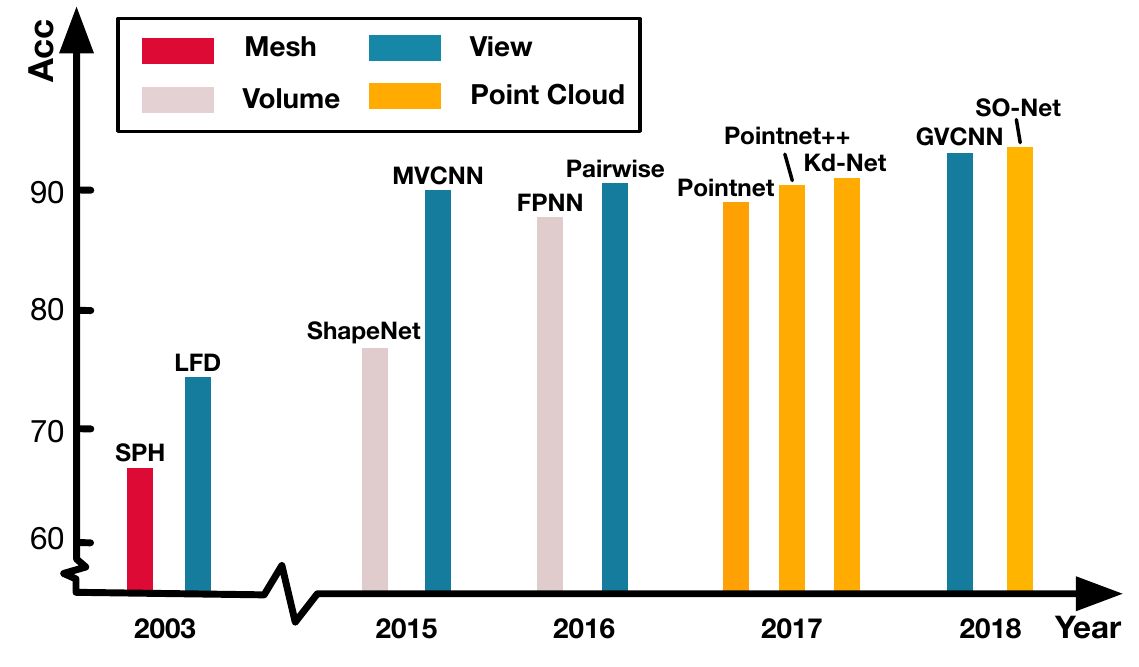}
    \caption{\textbf{The developing history of 3D shape representation using different types of data.} The X-axis indicates the proposed time of each method, and the Y-axis indicates the classification accuracy.}
    \label{fig:intro}
\end{figure}

\begin{figure*}[!htbp]
  \centering
  \includegraphics[width=6.5in]{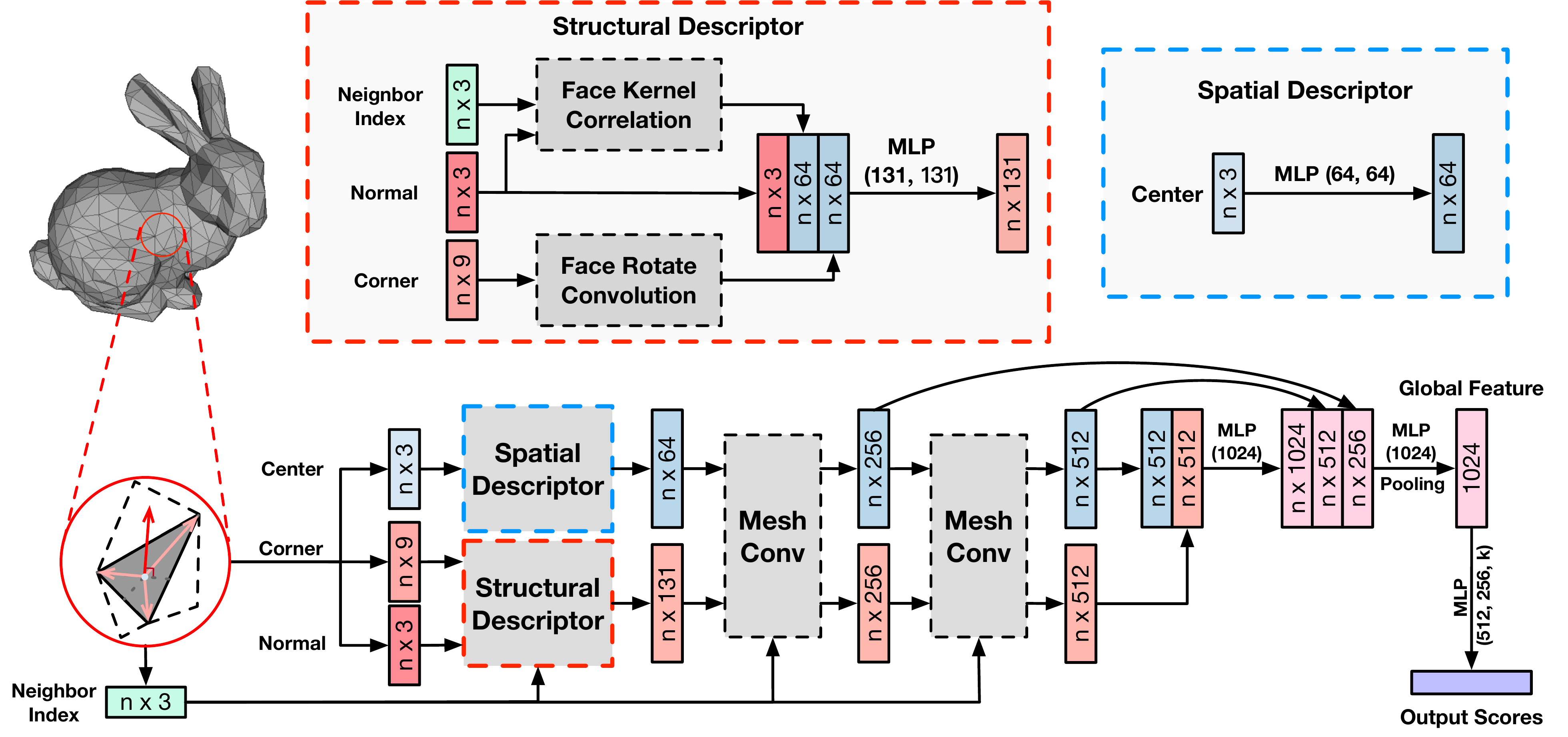}
  \caption{\textbf{The architecture of MeshNet.} The input is a list of faces with initial values, which are fed into the spatial and structural descriptors to generate initial spatial and structural features. The features are then aggregated with neighboring information in the mesh convolution blocks labeled as ``Mesh Conv'', and fed into a pooling function to output the global feature used for further tasks. Multi-layer-perceptron is labeled as ``mlp'' with the numbers in the parentheses indicating the dimension of the hidden layers and output layer.}
  \label{fig:pipeline}
\end{figure*}

Mesh data of 3D shapes is a collection of vertices, edges and faces, which is dominantly used in computer graphics for rendering and storing 3D models. Mesh data has the properties of complexity and irregularity. The complexity problem is that mesh consists of multiple elements, and different types of connections may be defined among them. The irregularity is another challenge for mesh data processing, which indicates that the number of elements in mesh may vary dramatically among 3D shapes, and permutations of them are arbitrary. In spite of these problems, mesh has stronger ability for 3D shape description than other types of data. Under such circumstances, how to effectively represent 3D shapes using mesh data is an urgent and challenging task.

In this paper, we present a mesh neural network, named MeshNet, that learns on mesh data directly for 3D shape representation. To deal with the challenges in mesh data processing, the faces are regarded as the unit and connections between faces sharing common edges are defined, which enables us to solve the complexity and irregularity problem with per-face processes and a symmetry function. Moreover, the feature of faces is split into spatial and structural features. Based on these ideas, we design the network architecture, with two blocks named spatial and structural descriptors for learning the initial features, and a mesh convolution block for aggregating neighboring features. In this way, the proposed method is able to solve the complexity and irregularity problem of mesh and represent 3D shapes well.

We apply our MeshNet method in the tasks of 3D shape classification and retrieval on the ModelNet40 \cite{wu20153d} dataset. And the experimental results show that MeshNet achieve significant improvement on 3D shape classification and retrieval using mesh data and comparable performance with recent methods using other types of 3D data.

The key contributions of our work are as follows:
\begin{itemize}
\item We propose a neural network using mesh for 3D shape representation and design blocks for capturing and aggregating features of polygon faces in 3D shapes.
\item We conduct extensive experiments to evaluate the performance of the proposed method, and the experimental results show that the proposed method performs well on the 3D shape classification and retrieval task.
\end{itemize}

\section{Related Work}
\subsection{Mesh Feature Extraction}
There are plenty of handcraft descriptors that extract features from mesh. \citeauthor{lien1984symbolic} calculate moments of each tetrahedron in mesh \cite{lien1984symbolic}, and \citeauthor{zhang2001efficient} develop more functions applied to each triangle and add all the resulting values as features \cite{zhang2001efficient}. \citeauthor{hubeli2001multiresolution} extend the features of surfaces to a multiresolution setting to solve the unstructured problem of mesh data \cite{hubeli2001multiresolution}. In SPH \cite{kazhdan2003rotation}, a rotation invariant representation is presented with existing orientation dependent descriptors. Mesh difference of Gaussians (DOG) introduces the Gaussian filtering to shape functions.\cite{zaharescu2009surface} Intrinsic shape context (ISC) descriptor \cite{kokkinos2012intrinsic} develops a generalization to surfaces and solves the problem of orientational ambiguity. 

\subsection{Deep Learning Methods for 3D Shape Representation}
With the construction of large-scale 3D model datasets, numerous deep descriptors of 3D shapes are proposed. Based on different types of data, these methods can be categorized into four types. 

\textit{Voxel-based method.} 3DShapeNets \cite{wu20153d} and VoxNet \cite{maturana2015voxnet} propose to learn on volumetric grids, which partition the space into regular cubes. However, they introduce extra computation cost due to the sparsity of data, which restricts them to be applied on more complex data. Field probing neural networks (FPNN) \cite{li2016fpnn}, Vote3D \cite{wang2015voting} and Octree-based convolutional neural network (OCNN) \cite{wang2017cnn} address the sparsity problem, while they are still restricted with input getting larger. 

\textit{View-based method.} Using 2D images of 3D shapes to represent them is proposed by Multi-view convolutional neural networks (MVCNN) \cite{su2015multi}, which aggregates 2D views from a loop around the object and applies 2D deep learning framework to them. Group-view convolutional neural networks (GVCNN) \cite{feng2018gvcnn} proposes a hierarchical framework, which divides views into different groups with different weights to generate a more discriminative descriptor for a 3D shape. This type of method also expensively adds the computation cost and is hard to be applied for tasks in larger scenes. 

\textit{Point-based method.} Due to the irregularity of data, point cloud is not suitable for previous frameworks. PointNet \cite{qi2017pointnet++} solves this problem with per-point processes and a symmetry function, while it ignores the local information of points. PointNet++ \cite{qi2017pointnet++} adds aggregation with neighbors to solve this problem. Self-organizing network (SO-Net) \cite{li2018so}, kernel correlation network (KCNet) \cite{shen2018mining} and PointSIFT \cite{jiang2018pointsift} develop more detailed approaches for capturing local structures with nearest neighbors. Kd-Net \cite{klokov2017escape} proposes another approach to solve the irregularity problem using k-d tree. 

\textit{Fusion method.} These methods learn on multiple types of data and fusion the features of them together. FusionNet \cite{hegde2016fusionnet} uses the volumetric grid and multi-view for classification. Point-view network (PVNet) \cite{you2018pvnet} proposes the embedding attention fusion to exploit both point cloud data and multi-view data.

\section{Method}
In this block, we present the design of MeshNet. Firstly, we analyze the properties of mesh, propose the methods for designing network and reorganize the input data. We then introduce the overall architecture of MeshNet and some blocks for capturing features of faces and aggregating them with neighbor information, which are then discussed in detail. 

\subsection{Overall Design of MeshNet}
We first introduce the mesh data and analyze its properties. Mesh data of 3D shapes is a collection of vertices, edges and faces, in which vertices are connected with edges and closed sets of edges form faces. In this paper, we only consider triangular faces. Mesh data is dominantly used for storing and rendering 3D models in computer graphics, because it provides an approximation of the smooth surfaces of objects and simplifies the rendering process. Numerous studies on 3D shapes in the field of computer graphic and geometric modeling are taken based on mesh.
\begin{figure}[!htbp]
    \centering
    \includegraphics[width=3.2in]{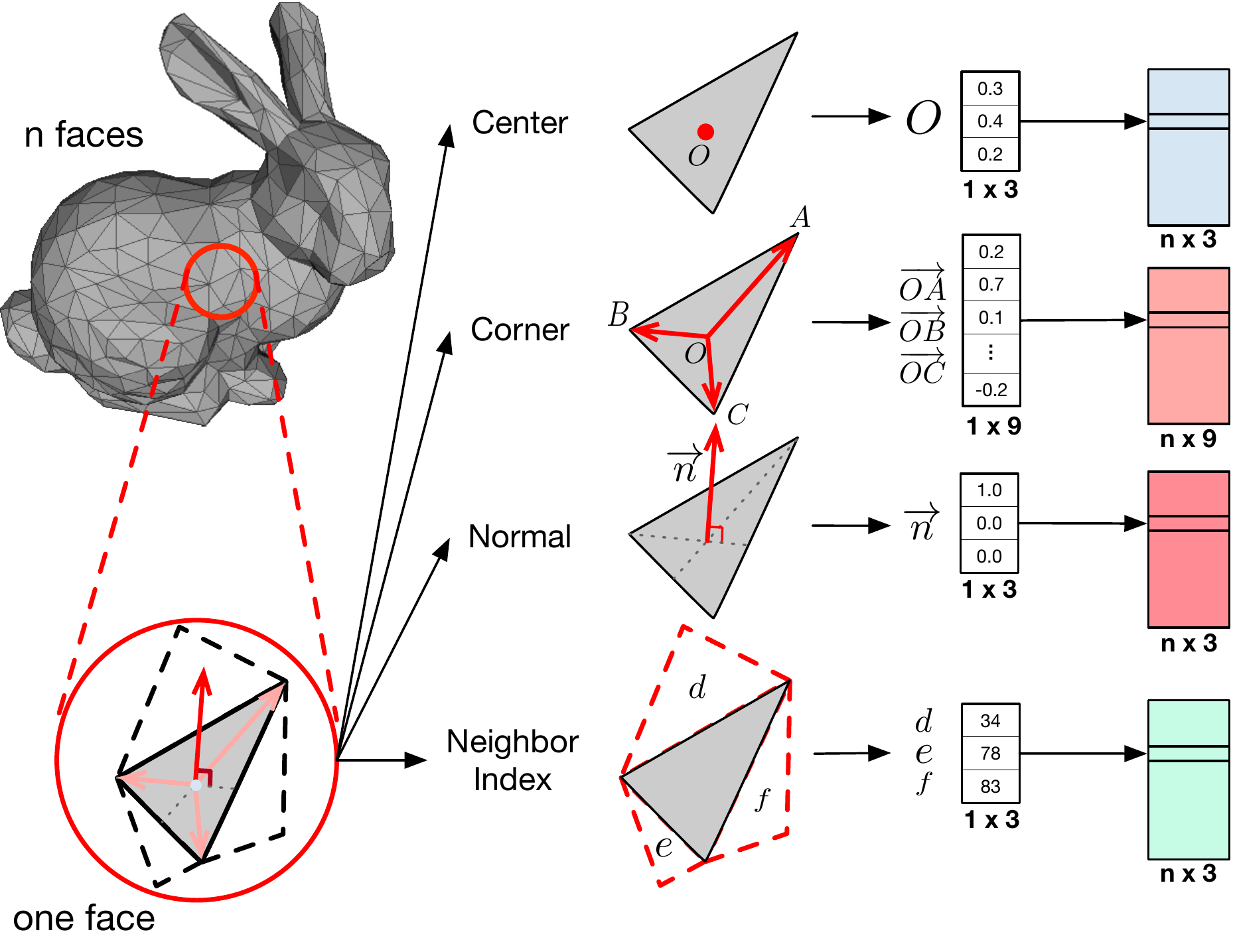}
    \caption{\textbf{Initial values of each face.} There are four types of initial values, divided into two parts: center, corner and normal are the face information, and neighbor index is the neighbor information.}
    \label{fig:input}
\end{figure}

Mesh data shows stronger ability to describe 3D shapes comparing with other popular types of data. Volumetric grid and multi-view are data types defined to avoid the irregularity of the native data such as mesh and point cloud, while they lose some natural information of the original object. For point cloud, there may be ambiguity caused by random sampling and the ambiguity is more obvious with fewer amount of points. In contrast, mesh is more clear and loses less natural information. Besides, when capturing local structures, most methods based on point cloud collect the nearest neighbors to approximately construct an adjacency matrix for further process,  while in mesh there are explicit connection relationships to show the local structure clearly.  However, mesh data is also more irregular and complex for the multiple compositions and varying numbers of elements.

\bigskip
To get full use of the advantages of mesh and solve the problem of its irregularity and complexity, we propose two key ideas of design:
\begin{itemize}
\item \textbf{Regard face as the unit.} Mesh data consists of multiple elements and connections may be defined among them. To simplify the data organization, we regard face as the only unit and define a connection between two faces if they share a common edge. There are several advantages of this simplification. First is that one triangular face can connect with no more than three faces, which makes the connection relationship regular and easy to use. More importantly, we can solve the disorder problem with per-face processes and a symmetry function, which is similar to PointNet \cite{qi2017pointnet}, with per-face processes and a symmetry function. And intuitively, face also contains more information than vertex and edge.
\item \textbf{Split feature of face.} Though the above simplification enables us to consume mesh data similar to point-based methods, there are still some differences between point-unit and face-unit because face contains more information than point. We only need to know ``where you are" for a point, while we also want to know ``what you look like" for a face. Correspondingly, we split the feature of faces into \textbf{spatial feature} and \textbf{structural feature}, which helps us to capture features more explicitly.
\end{itemize}

\bigskip
Following the above ideas, we transform the mesh data into a list of faces. For each face, we define the initial values of it, which are divided into two parts (illustrated in Fig \ref{fig:input}):
\begin{itemize}
\item Face Information:
\begin{itemize}
\item \textbf{Center}: coordinate of the center point
\item \textbf{Corner}: vectors from the center point to three vertices
\item \textbf{Normal}: the unit normal vector
\end{itemize}
\item Neighbor Information:
\begin{itemize}
\item \textbf{Neighbor Index}: indexes of the connected faces (filled with the index of itself if the face connects with less than three faces)
\end{itemize}
\end{itemize}

\bigskip
In the final of this section, we present the overall architecture of MeshNet, illustrated in Fig \ref{fig:pipeline}. A list of faces with initial values is fed into two blocks, named \textbf{spatial descriptor} and \textbf{structural descriptor}, to generate the initial spatial and structural features of faces. The features are then passed through some \textbf{mesh convolution} blocks to aggregate neighboring information, which get features of two types as input and output new features of them. It is noted that all the processes above work on each face respectively and share the same parameters. After these processes, a pooling function is applied to features of all faces for generating global feature, which is used for further tasks. The above blocks will be discussed in following sections.

\subsection{Spatial and Structural Descriptors}
We split feature of faces into spatial feature and structural feature. The spatial feature is expected to be relevant to the spatial position of faces, and the structural feature is relevant to the shape information and local structures. In this section, we present the design of two blocks, named spatial and structural descriptors, for generating initial spatial and structural features.

\paragraph{Spatial descriptor}
The only input value relevant to spatial position is the center value. In this block, we simply apply a shared MLP to each face's center, similar to the methods based on point cloud, and output initial spatial feature.

\paragraph{Structural descriptor: face rotate convolution}
We propose two types of structural descriptor, and the first one is named face rotate convolution, which captures the ``inner'' structure of faces and focus on the shape information of faces. The input of this block is the corner value.

\begin{figure}[!tbp]
    \centering
    \includegraphics[width=3.0in]{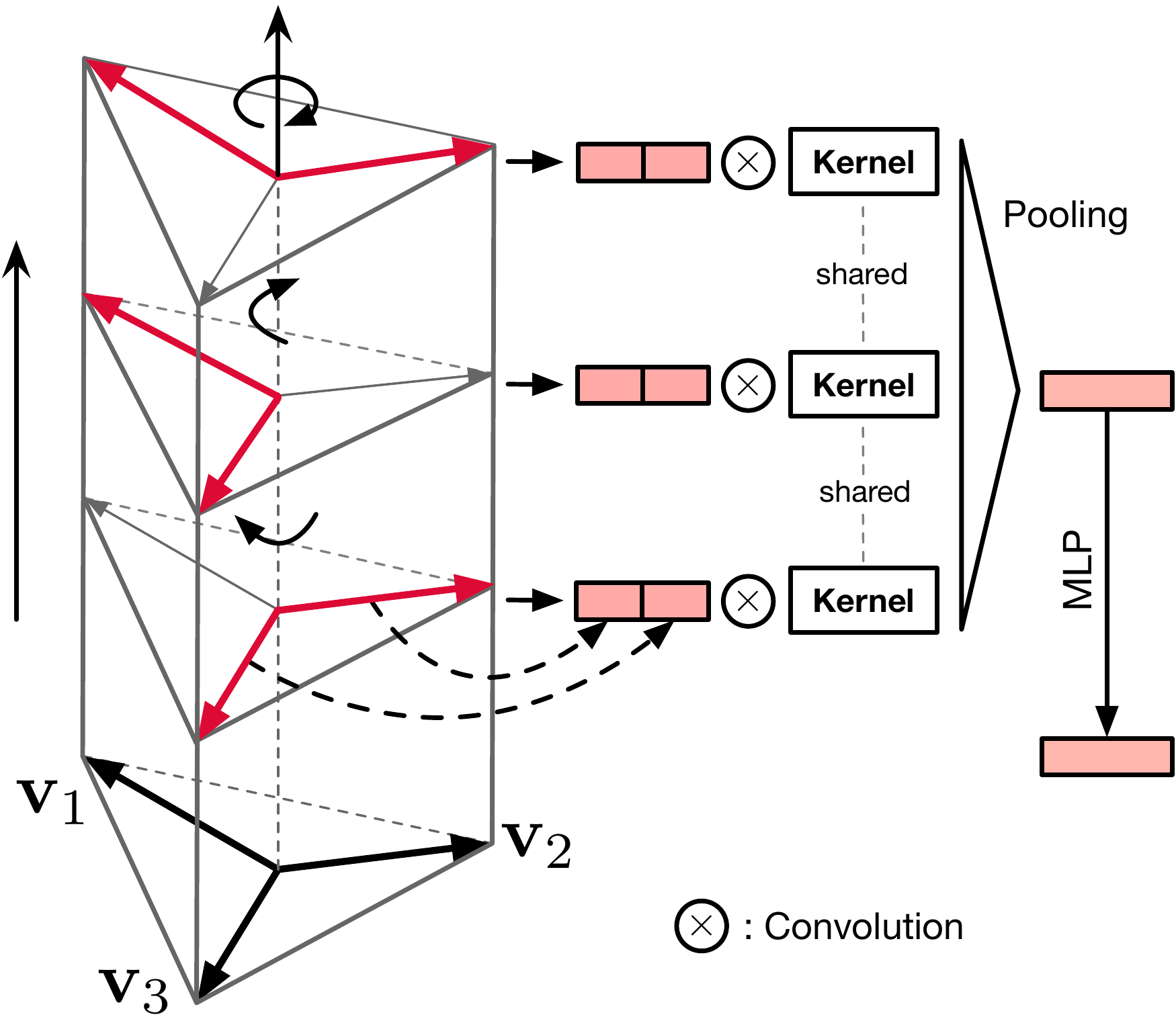}
    \caption{\textbf{The face rotate convolution block.} Kernels rotate through the face and are applied to pairs of corner vectors for the convolution operation.}
    \label{fig:rc}
\end{figure}

The operation of this block is illustrated in Fig \ref{fig:rc}. Suppose the corner vectors of a face are \(\mathbf{v}_1, \mathbf{v}_2,  \mathbf{v}_3\), we define the output value of this block as follows: 
\begin{equation}
	g(\frac{1}{3}(f(\mathbf{v}_1, \mathbf{v}_2) + f(\mathbf{v}_2, \mathbf{v}_3) + f(\mathbf{v}_3, \mathbf{v}_1))),
\end{equation}
where \(f(\cdot, \cdot) : \mathbb{R}^3 \times \mathbb{R}^3 \rightarrow \mathbb{R}^{K_1}\) and \(g(\cdot) : \mathbb{R}^{K_1} \rightarrow \mathbb{R}^{K_2}\) are any valid functions.

This process is similar to a convolution operation, with two vectors as the kernel size, one vector as the stride and \(K_1\) as the number of kernels, except that translation of kernels is replaced by rotation. The kernels, represented by \(f(\cdot, \cdot)\), rotates through the face and works on two vectors each time. 

With the above process, we eliminate the influence caused by the order of processing corners, avoid individually considering each corner and also leave full space for mining features inside faces. After the rotate convolution, we apply an average pooling and a shared MLP as \(g(\cdot)\) to each face, and output features with the length of \(K_2\).

\paragraph{Structural descriptor: face kernel correlation} 
Another structural descriptor we design is the face kernel correlation, aiming to capture the ``outer'' structure of faces and explore the environments where faces locate. The method is inspired by KCNet \cite{shen2018mining}, which uses kernel correlation (KC) \cite{tsin2004correlation} for mining local structures in point clouds. KCNet learns kernels representing different spatial distributions of point sets, and measures the geometric affinities between kernels and neighboring points for each point to indicate the local structure. However, this method is also restricted by the ambiguity of point cloud, and may achieve better performance in mesh.

In our face kernel correlation, we select the normal values of each face and its neighbors as the source, and learnable sets of vectors as the reference kernels. Since all the normals we use are unit vectors, we model vectors of kernels with parameters in the spherical coordinate system, and parameters \((\theta, \phi)\) represent the unit vector \((x, y, z)\) in the Euclidean coordinate system:
\begin{equation}
\left\{
             \begin{array}{lr}
             x=\sin\theta\cos\phi  \\
             y=\sin\theta\sin\phi \\
             z=\cos\theta
             \end{array},
\right.
\end{equation}
where \(\theta \in [0, \pi]\) and \(\phi \in [0, 2\pi)\).

We define the kernel correlation between the i-th face and the k-th kernel as follows: 
\begin{equation}
    KC(i, k) = \frac{1}{|\mathcal{N}_i||\mathcal{M}_k|}\sum\limits_{\mathbf{n} \in \mathcal{N}_i}\sum\limits_{\mathbf{m} \in \mathcal{M}_k}K_{\sigma}(\mathbf{n}, \mathbf{m}),
\end{equation}
where \(\mathcal{N}_i\) is the set of normals of the i-th face and its neighbor faces, \(\mathcal{M}_k\) is the set of normals in the k-th kernel, and \(K_{\sigma}(\cdot,\cdot)\ : \mathbb{R}^3 \times \mathbb{R}^3 \rightarrow \mathbb{R}\) is the kernel function indicating the affinity between two vectors. In this paper, we generally choose the Gaussian kernel:
\begin{equation}
    K_{\sigma}(\mathbf{n}, \mathbf{m}) = \exp(-\frac{\left\| \mathbf{n} - \mathbf{m}\right \|^2}{2\sigma^2}),
\end{equation}
where \(\left\|\cdot\right\|\) is the length of a vector in the Euclidean space, and \(\sigma\) is the hyper-parameter that controls the kernels' resolving ability or tolerance to the varying of sources. 

With the above definition, we calculate the kernel correlation between each face and kernel, and more similar pairs will get higher values. Since the parameters of kernels are learnable, they will turn to some common distributions on the surfaces of 3D shapes and be able to describe the local structures of faces. We set the value of \(KC(i,k)\) as the k-th feature of the i-th face. Therefore, with \(M\) learnable kernels, we generate features with the length of \(M\) for each face.

\subsection{Mesh Convolution}
The mesh convolution block is designed to expand the receptive field of faces, which denotes the number of faces perceived by each face, by aggregating information of neighboring faces. In this process, features related to spatial positions should not be included directly because we focus on faces in a local area and should not be influenced by where the area locates. In the 2D convolutional neural network, both the convolution and pooling operations do not introduce any positional information directly while aggregating with neighboring pixels' features. Since we have taken out the structural feature that is irrelevant to positions, we only aggregate them in this block. 

\begin{figure}[!htbp]
  \centering
  \includegraphics[width=3.5in]{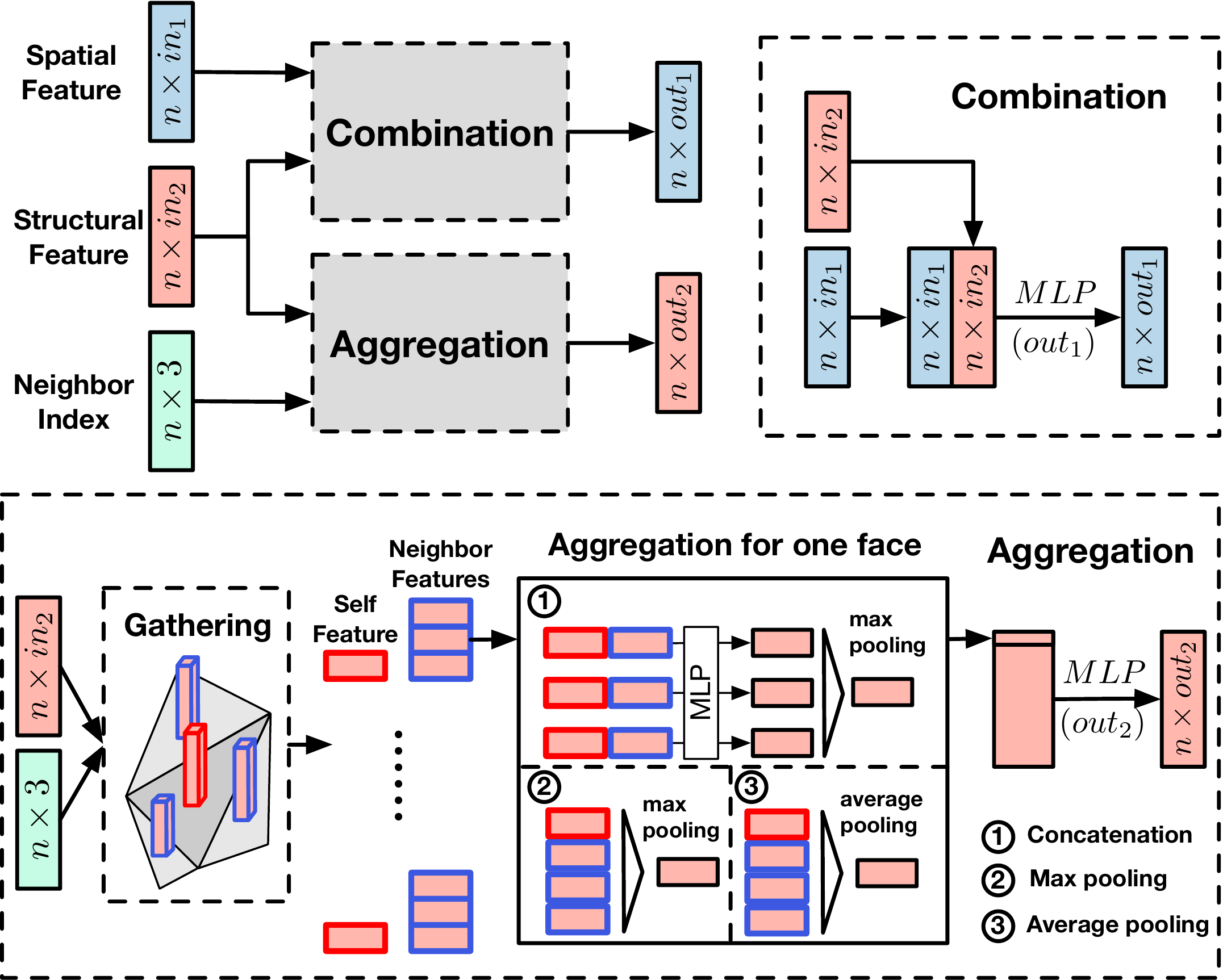}
  \caption{\textbf{The mesh convolution.} ``Combination'' donates the combination of spatial and structural feature. ``Aggregation'' denotes the aggregation of structural feature, in which ``Gathering'' denotes the process of getting neighbors' features and ``Aggregation for one face'' denotes different methods of aggregating features.}
  \label{fig:meshconv}
\end{figure}

Aggregation of structural feature enables us to capture structures of wider field around each face. Furthermore, to get more comprehensive feature, we also combine the spatial and structural feature together in this block. The mesh convolution block contains two parts: \textbf{combination of spatial and structural features} and \textbf{aggregation of structural feature}, which respectively output the new spatial and structural features. Fig \ref{fig:meshconv} illustrates the design of this block.

\paragraph{Combination of Spatial and Structural Features}
We use one of the most common methods of combining two types of features, which concatenates them together and applies a MLP. As we have mentioned, the combination result, as the new spatial feature, is actually more comprehensive and contains both spatial and structural information. Therefore, in the pipeline of our network, we concatenate all the combination results for generating global feature.

\paragraph{Aggregation of Structural Feature}
With the input structural feature and neighbor index, we aggregate the feature of a face with features of its connected faces. Several aggregation methods are listed and discussed as follows:
\begin{itemize}
\item \textbf{Average pooling: } The average pooling may be the most common aggregation method, which simply calculates the average value of features in each channel. However, this method sometimes weakens the strongly-activated areas of the original features and reduce the distinction of them.
\item \textbf{Max pooling: } Another pooling method is the max pooling, which calculates the max value of features in each channel. Max pooling is widely used in 2D and 3D deep learning frameworks for its advantage of maintaining the strong activation of neurons.
\item \textbf{Concatenation: } We define the concatenation aggregation, which concatenates the feature of face with feature of each neighbor respectively, passes these pairs through a shared MLP and applies a max pooling to the results. This method both keeps the original activation and leaves space for the network to combine neighboring features.
\end{itemize}

We finally use the concatenation method in this paper. After aggregation, another MLP is applied to further fusion the neighboring features and output new structural feature.

\subsection{Implementation Details}
Now we present the details of implementing MeshNet, illustrated in Fig 2, including the settings of hyper-parameters and some details of the overall architecture. 

The spatial descriptor contains fully-connected layers (64, 64) and output a initial spatial feature with length of 64. Parameters inside parentheses indicate the dimensions of layers except the input layer. In the face rotate convolution, we set \(K_1=32\) and \(K_2 = 64\), and correspondingly, the functions \(f(\cdot, \cdot)\) and \(g(\cdot)\) are implemented by fully-connected layers (32, 32) and (64, 64). In the face kernel correlation, we set \(M = 64\) (64 kernels) and \(\sigma = 0.2\).

We parameterize the mesh convolution block with a four-tuple \((in_1, in_2, out_1, out_2)\), where \(in_1\) and \(out1\) indicate the input and output channels of spatial feature, and the  \(in_2\) and \(out2\) indicates the same of structural feature. The two mesh convolution blocks used in the pipeline of MeshNet are configured as \((64, 131, 256, 256)\) and \((256, 256, 512, 512)\).

\section{Experiments}
In the experiments, we first apply our network for 3D shape classification and retrieval. Then we conduct detailed ablation experiments to analyze the effectiveness of blocks in the architecture. We also investigate the robustness to the number of faces and the time and space complexity of our network. Finally, we visualize the structural features from the two structural descriptors.

\subsection{3D Shape Classification and Retrieval}
We apply our network on the ModelNet40 dataset \cite{wu20153d} for classification and retrieval tasks. The dataset contains 12,311 mesh models from 40 categories, in which 9,843 models for training and 2,468 models for testing. For each model, we simplify the mesh data into no more than 1,024 faces, translate it to the geometric center, and normalize it into a unit sphere. Moreover, we also compute the normal vector and indexes of connected faces for each face. During the training, we augment the data by jittering the positions of vertices by a Gaussian noise with zero mean and 0.01 standard deviation. Since the number of faces varies among models, we randomly fill the list of faces to the length of 1024 with existing faces for batch training.
\begin{table}[!bp]
    \centering
    \begin{tabular}{p{1.7in}|c|c|c}
        \toprule
         \multirow{2}{*}{Method} & \multirow{2}{*}{Modality} & Acc & mAP \\
         & & (\%) & (\%) \\
         \midrule
         3DShapeNets \cite{wu20153d} & volume & 77.3 & 49.2 \\
         VoxNet \cite{maturana2015voxnet} & volume & 83.0 & - \\
         FPNN \cite{li2016fpnn} & volume & 88.4 & - \\
         \midrule
         LFD \cite{chen2003visual} & view & 75.5 & 40.9 \\
         MVCNN \cite{su2015multi} & view & 90.1 & 79.5 \\
         Pairwise \cite{johns2016pairwise} & view & 90.7 & - \\
         \midrule
         PointNet \cite{qi2017pointnet} & point & 89.2 & - \\
         PointNet++ \cite{qi2017pointnet++} & point & 90.7 & - \\
         Kd-Net \cite{klokov2017escape} & point & 91.8 & - \\
         KCNet \cite{shen2018mining} & point & 91.0 & - \\
         \midrule
         SPH \cite{kazhdan2003rotation} & mesh & 68.2 & 33.3 \\
         MeshNet & mesh & 91.9 & 81.9 \\
         \bottomrule
    \end{tabular}
    \caption{Classification and retrieval results on ModelNet40.}
    \label{tab:application}
\end{table}

For classification, we apply fully-connected layers (512, 256, 40) to the global features as the classifier, and add dropout layers with drop probability of 0.5 before the last two fully-connected layers. For retrieval, we calculate the L2 distances between the global features as similarities and evaluate the result with mean average precision (mAP). We use the SGD optimizer for training, with initial learning rate 0.01, momentum 0.9, weight decay 0.0005 and batch size 64.

\begin{table}[!tbp]
    \centering
    \begin{tabular}{l|c|c|c|c|c|c}
        \toprule
         Spatial &     & \(\checkmark\) & \(\checkmark\) & \(\checkmark\) &  \(\checkmark\) &  \(\checkmark\)\\
         Structural-FRC & \(\checkmark\) &     &      & \(\checkmark\) &  \(\checkmark\) &  \(\checkmark\)\\
         Structural-FKC & \(\checkmark\) &     & \(\checkmark\) &      &  \(\checkmark\) &  \(\checkmark\)\\
         Mesh Conv & \(\checkmark\) & \(\checkmark\) & \(\checkmark\) & \(\checkmark\) &  &  \(\checkmark\)\\
         \midrule
         Accuracy (\%) & 83.5 & 88.2 & 87.0 & 89.9 & 90.4 & 91.9\\
         \bottomrule
    \end{tabular}
    \caption{Classification results of ablation experiments on ModelNet40.}
    \label{tab:ablation}
\end{table}

\begin{table}[!tbp]
    \centering
    \begin{tabular}{l|c}
        \toprule
         Aggregation Method & Accuracy (\%)\\
         \midrule
         Average Pooling & 90.7 \\
         Max Pooling & 91.1 \\
         Concatenation & 91.9 \\
         \bottomrule
    \end{tabular}
    \caption{Classification results of different aggregation methods on ModelNet40.}
    \label{tab:aggregation}
\end{table}

Table \ref{tab:application} shows the experimental results of classification and retrieval on ModelNet40, comparing our work with representative methods. It is shown that, as a mesh-based representation, MeshNet achieves satisfying performance and makes great improvement compared with traditional mesh-based methods. It is also comparable with recent deep learning methods based on other types of data.

Our performance is dedicated to the following reasons. With face-unit and per-face processes, MeshNet solves the complexity and irregularity problem of mesh data and makes it suitable for deep learning method. Though with deep learning's strong ability to capture features, we do not simply apply it, but design blocks to get full use of the rich information in mesh. Splitting features into spatial and structural features enables us to consider the spatial distribution and local structure of shapes respectively. And the mesh convolution blocks widen the receptive field of faces. Therefore, the proposed method is able to capture detailed features of faces and conduct the 3D shape representation well.

\subsection{On the Effectiveness of Blocks}
To analyze the design of blocks in our architecture and prove the effectiveness of them, we conduct several ablation experiments, which compare the classification results while varying the settings of architecture or removing some blocks. 

For the spatial descriptor, labeled as ``Spatial'' in Table \ref{tab:ablation}, we remove it together with the use of spatial feature in the network, and maintain the aggregation of structural feature in the mesh convolution.

For the structural descriptor,  we first remove the whole of it and use max pooling to aggregate the spatial feature in the mesh convolution. Then we partly remove the face rotate convolution,  labeled as ``Structural-FRC'' in Table \ref{tab:ablation},  or the face kernel correlation, labeled as ``Structural-FKC'', and keep the rest of pipeline to prove the effectiveness of each structural descriptor.

For the mesh convolution, labeled as ``Mesh Conv'' in Table \ref{tab:ablation}, we remove it and use the initial features to generate the global feature directly. We also explore the effectiveness of different aggregation methods in this block, and compare them in Table \ref{tab:aggregation}. The experimental results show that the adopted concatenation method performs better for aggregating neighboring features.

\subsection{On the Number of Faces}
The number of faces in ModelNet40 varies dramatically among models. To explore the robustness of MeshNet to the number of faces, we regroup the test data by the number of faces with interval 200. In Table \ref{tab:facenum}, we list the proportion of the number of models in each group, together with the classification results. It is shown that the accuracy is absolutely irrelevant to the number of faces and shows no downtrend with the decrease of it, which proves the great robustness of MeshNet to the number of faces. Specifically, on the 9 models with less than 50 faces (the minimum is 10), our network achieves 100\% classification accuracy, showing the ability to represent models with extremely few faces. 

\begin{table}[!htbp]
    \centering
    \begin{tabular}{c |c|c}
        \toprule
         Number of Faces & Proportion (\%) & Accuracy (\%)\\
         \midrule
         \([ 1000, 1024)\) & 69.48 & 92.00 \\
         \([800, 1000)\) & 6.90 & 92.35 \\
         \([600, 800)\) & 4.70 & 93.10 \\
         \([400, 600)\) & 6.90 & 91.76 \\
         \([200, 400)\) & 6.17 & 90.79 \\
         \([0, 200)\) & 5.84 & 90.97 \\
         \bottomrule
    \end{tabular}
    \caption{Classification results of groups with different number of faces on ModelNet40.}
    \label{tab:facenum}
\end{table}

\subsection{On the Time and Space Complexity}
Table \ref{tab:timespace} compares the time and space complexity of our network with several representative methods based on other types of data for the classification task. The column labeled ``\#params'' shows the total number of parameters in the network and the column labeled ``FLOPs/sample'' shows the number of floating-operations conducted for each input sample, representing the space and time complexity respectively.

It is known that methods based on volumetric grid and multi-view introduce extra computation cost while methods based on point cloud are more efficient. Theoretically, our method works with per-face processes and has a linear complexity to the number of faces. In Table \ref{tab:timespace}, MeshNet shows comparable effectiveness with methods based on point cloud in both time and space complexity, leaving enough space for further development.

\begin{table}[!htbp]
    \centering
    \begin{tabular}{l|c|c}
        \toprule
         \multirow{2}{*}{Method} & \#params & FLOPs /\\
         & (M) & sample (M)\\
         \midrule
         PointNet \cite{qi2017pointnet} & 3.5 & 440 \\
         Subvolume \cite{qi2016volumetric} & 16.6 & 3633 \\
         MVCNN \cite{su2015multi} & 60.0 & 62057 \\
         \midrule
         MeshNet & 4.25 & 509 \\
         \bottomrule
    \end{tabular}
    \caption{Time and space complexity for classification.}
    \label{tab:timespace}
\end{table}

\begin{figure}[!tbp]
    \centering
    \includegraphics[width=3.3 in]{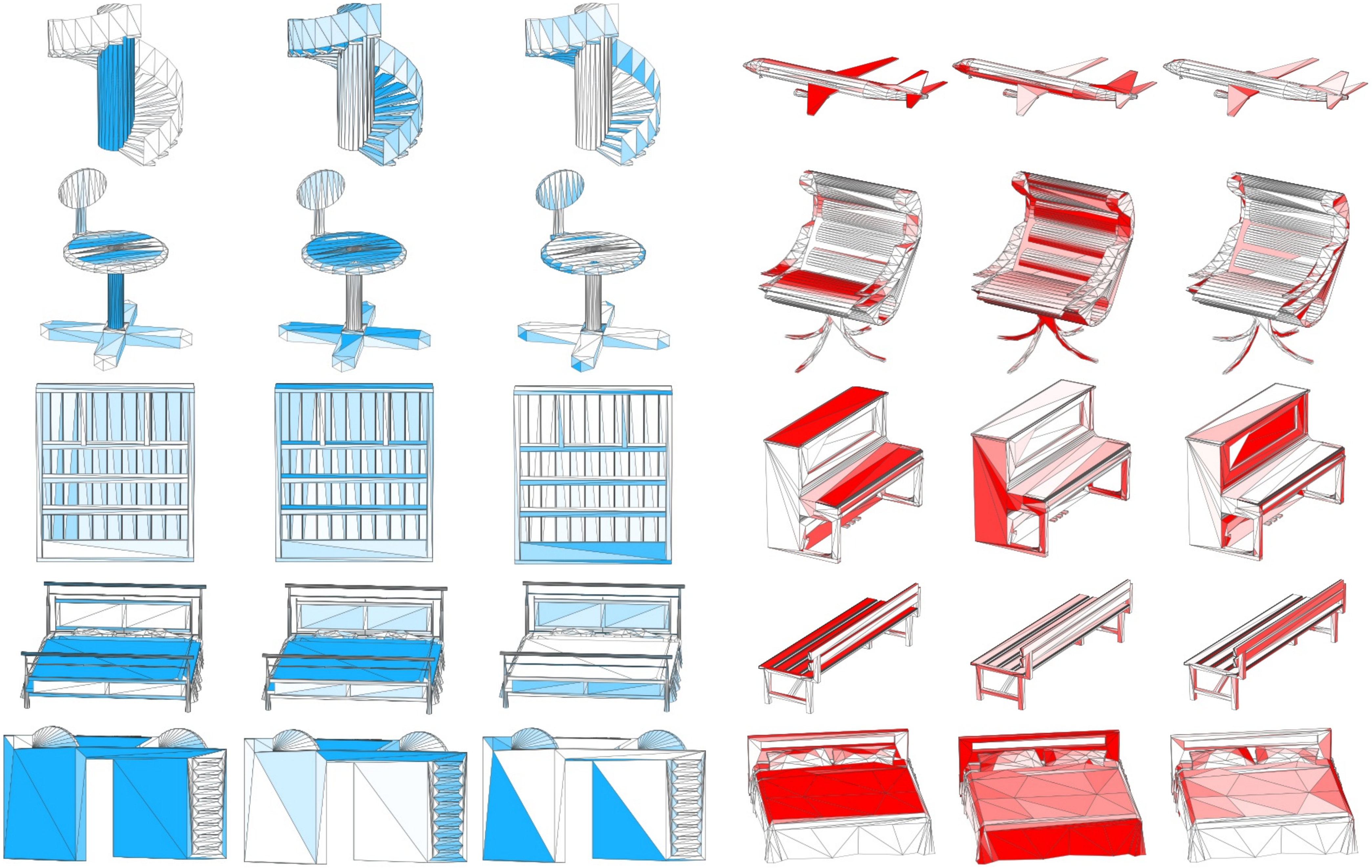}
    \caption{\textbf{Feature visualization of structural feature.} Models from the same column are colored with their values of the same channel in features. \textbf{Left}: Features from the face rotate convolution. \textbf{Right}: Features from the face kernel correlation.}
    \label{fig:vis}
\end{figure}
\subsection{Feature Visualization}
To figure out whether the structural descriptors successfully capture our features of faces as expected, we visualize the two types of structural features from face rotate convolution and face kernel correlation. We randomly choose several channels of these features, and for each channel, we paint faces with colors of different depth corresponding to their values in this channel. 

The left of Fig \ref{fig:vis} visualizes features from face rotate convolution, which is expected to capture the ``inner'' features of faces and concern about the shapes of them. It is clearly shown that these faces with similar look are also colored similarly, and different channels may be activated by different types of triangular faces.

The visualization results of features from face kernel correlation are in the right of Fig \ref{fig:vis}. As we have mentioned, this descriptor captures the ``outer'' features of each face and is relevant to the whole appearance of the area where the face locates. In the visualization, faces in similar types of areas, such as flat surfaces and steep slopes, turn to have similar features, regardless of their own shapes and sizes.

\section{Conclusions}
In this paper, we propose a mesh neural network, named MeshNet, which learns on mesh data directly for 3D shape representation. The proposed method is able to solve the complexity and irregularity problem of mesh data and conduct 3D shape representation well. In this method, the polygon faces are regarded as the unit and features of them are split into spatial and structural features. We also design blocks for capturing and aggregating features of faces. We conduct experiments for 3D shape classification and retrieval and compare our method with the state-of-the-art methods. The experimental result and comparisons demonstrate the effectiveness of the proposed method on 3D shape representation. In the future, the network can be further developed for more computer vision tasks.

\section{Acknowledgments}
This work was supported by National Key R\&D Program of China (Grant No. 2017YFC0113000), National Natural Science Funds of China (U1701262, 61671267), National Science and Technology Major Project (No. 2016ZX01038101), MIIT IT funds (Research and application of TCN key technologies) of China, and The National Key Technology R\&D Program (No. 2015BAG14B01-02).

\bibliographystyle{aaai}

\end{document}